\pdfoutput=1

\documentclass[11pt]{article}

\usepackage{emnlp2021}

\usepackage{times}
\usepackage{latexsym}
\usepackage{multirow}
\usepackage{makecell}

\usepackage[T1]{fontenc}

\usepackage[utf8]{inputenc}

\usepackage{microtype}

\newcommand{\avocado}[0]{\includegraphics[width=.6cm]{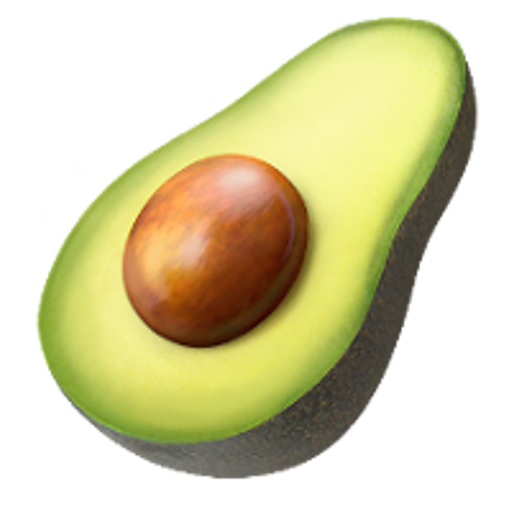}}

\usepackage{latexsym}
\usepackage{booktabs}
\usepackage{url}
\usepackage{graphicx}
\usepackage{comment}
\usepackage{amssymb}
\usepackage{caption}
\usepackage{subcaption}
\usepackage{enumitem}

\newcommand{\changed}[1]{{#1}}

\newcommand{\changedD}[1]{#1}

\usepackage{adjustbox}
\usepackage{array}

\newcolumntype{R}[2]{%
    >{\adjustbox{angle=#1,lap=\width-(#2)}\bgroup}%
    l%
    <{\egroup}%
}
\newcommand*\rot{\multicolumn{1}{R{40}{3em}}}

\providecommand{\hanna}[1]{
    {\protect\color{blue}{[Hanna: #1]}}
}
\newcommand{\egbox}[1]{
\smallskip
\noindent
\fbox{
        \parbox{0.95\linewidth}{
        \vspace{0.5ex}#1\vspace{0.5ex}}
      }
\smallskip
}

\definecolor{darkgreen}{rgb}{0.0, 0.5, 0.0}

\newcommand{\name}{\textsc{GooAQ}}
\title{\name: 3 Million+ Question-Answers from Google }
\title{\name: 3 Million+ Question-Answers Extracted from Google}
\title{\name: 3 Million+ Natural Questions with Diverse Answers}
\title{\name: Large-scale Set of Natural Open-Domain Questions \\  with Diverse Answers }
\title{\name: Large-scale Set of Natural Open-Domain Questions \\  with Diverse Answers }
\title{Is Your QA System as Good as Google's?}
\title{Is Your QA System as Good as Google?}
\title{Is Your Open-Domain QA System as Good as Google's?}
\title{Is Your Language-Model as Good as Google QA?}
\title{Is Your Language-Model as Knowledgeable as Google QA?}
\title{Is Your Language Model as Knowledgeable as Google?}
\title{\name: A Study of Language Models in Addressing \\ 
Open-Domain Questions of Various Answer Types}
\title{\name: An Analysis of Open-Response Question Answering \\ 
with Diverse Answer Types}
\title{\name: Open-Response Question Answering with Diverse Answer Types}
\title{
\vspace*{-0.5in}
    {{\small \hfill EMNLP-Findings'21}\\
    \vspace*{.25in}} 
\name \avocado: Open Question Answering with Diverse Answer Types
}

\newcommand{\task}[1]{$\mathcal{T}_{#1}$}

\author{
Daniel Khashabi$^{1}$ $\;\;\;$ Amos Ng   \\ 
\textbf{Tushar Khot}$^{1}$ $\;$ \textbf{Ashish Sabharwal}$^{1}$ $\;$ \textbf{Hannaneh Hajishirzi}$^{1,2}$ $\;$ \textbf{Chris Callison-Burch}$^{3}$
\\\\
 $^1$Allen Institute for AI, Seattle, WA, USA \\
 $^2$University of Washington, Seattle, WA, USA \\
 $^3$University of Pennsylvania, Philadelphia, PA, USA
}

\newcommand{\subscript}[2]{$#1 _ #2$}

\begin{document}

\maketitle

\begin{abstract} 
While day-to-day questions come with a variety of answer types, the current \changed{open} question-answering (QA) literature
\changed{represents isolated efforts on niche response types, with a heavy focus on specific kinds of short responses (people, places, etc.).}
\changed{To address this gap,}
we present \name, a large-scale dataset collected from Google questions and answers, containing 3 million questions with diverse answer types ranging from factual short answers to snippets to collections. Our human evaluation shows that 94\% of the  mined answers are accurate, enabling fine-tuning a pre-trained language model for answering \name{} questions. 
We use this dataset to study inherent differences between models \changed{producing different answer types, and observe interesting trends}.
\changed{For example,} in line with recent work, LM's strong performance on \name's short-answer questions heavily benefits from annotated data. However, their \changed{surprisingly high} quality in generating coherent and accurate answers for questions requiring long responses (such as `how' and `why' questions) is less reliant on observing annotated data and mainly supported by their pre-training. Moreover, we show that \name{} \changed{is a valuable training resource, resulting in strong performance on the recent ELI5 long-answers dataset.} We release \name{} to facilitate further research on improving QA with diverse response types.\footnote{\label{footnote:data-url}The dataset is available at \url{https://github.com/allenai/gooaq} under an appropriate license. 
}
\end{abstract}

\begin{figure*}
    \centering
    \includegraphics[scale=0.61, trim=0.1cm 0.1cm 0cm -0.1cm, clip=false]{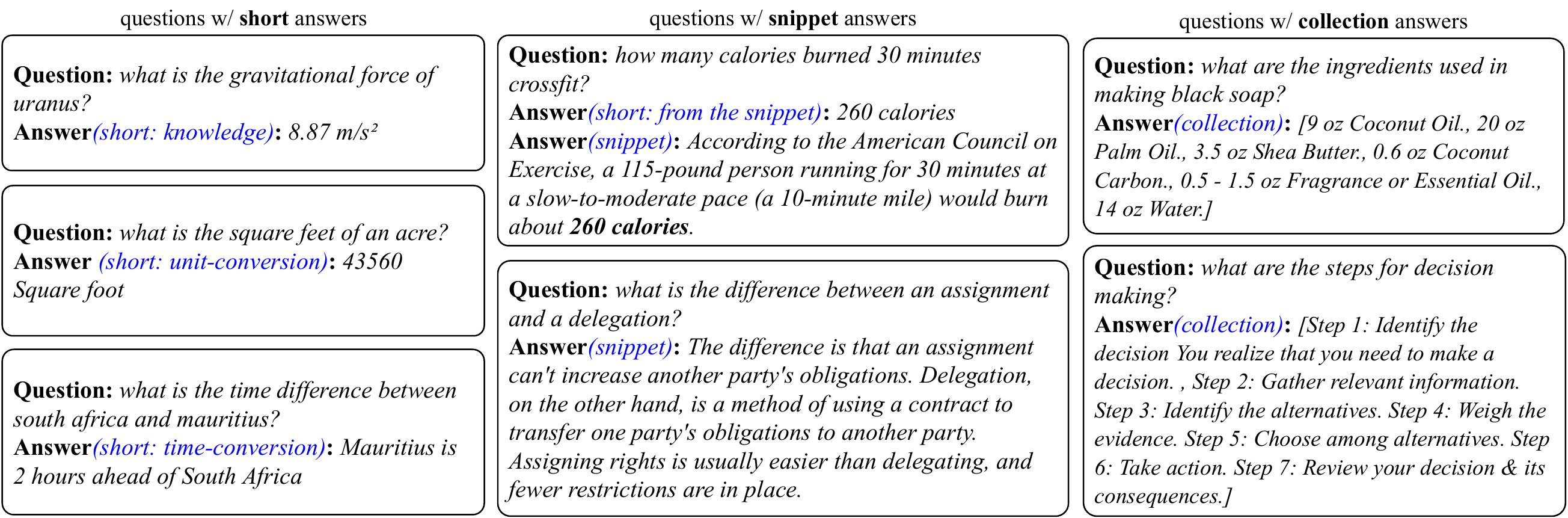}
    \caption{
        Examples from \name{} showing different types of the questions considered in this study. Each input is a  natural language question, mapped to  textual answer(s). 
        The questions/answers come with answer {\color{blue} type} which are inferred from meta information of the search results. 
    }
    \label{fig:examples}
\end{figure*}

\section{Introduction}

Research in ``open'' question answering (also referred to as open-response, open-domain, or direct answer QA) has resulted in numerous datasets and powerful models for answering questions without a specified context. This task requires the use of background knowledge either stored in the QA model or retrieved from large corpora or knowledge bases~\cite{roberts2020much,lewis2020question}.
Existing effort, however, involves isolated studies on niche answer types, mainly short responses and, in a few cases, long responses~\cite{joshi2017triviaqa,lee2019latent,bhakthavatsalam2021think}.

In contrast, many of the everyday questions that humans deal with and pose to search engines have a more diverse set of response types, as illustrated in Fig.~\ref{fig:examples}. Their answer can be a multi-sentence description (a \emph{snippet}) (e.g., `what is' or `can you' questions), a \emph{collection} of items such as ingredients (`what are kinds of', `things to') or of steps towards a goal such as unlocking a phone (`how to'), etc. Even when the answer is short, it can have rich types, e.g., unit conversion, time zone conversion, or a variety of knowledge look-up (`how much', `when is', etc.).\footnote{In contrast, the short responses in existing datasets typically inquire about people, dates, and counts. For instance, 65\% of Natural Questions~\cite{kwiatkowski2019natural} begin with `who', `when', or `how many'; cf.~Fig~\ref{fig:bigram:distributions}.} Such answer type diversity is not represented in any existing dataset. 

Motivated by this, we introduce \name\ (pronounced \textit{guac} like \textit{guacamole}), the first open QA benchmark containing \emph{questions with all of the above answer types within a unified dataset, collected using the same, coherent process}. \name\ contains $3$ million questions with short, snippet, or collection answers, such as the ones shown in Fig.~\ref{fig:examples}.
Besides supporting research on various types of answers, \name\ enables a quantitative study of the inherent differences in systems across different answer types.

\name{} questions are automatically mined from Google's search-autocomplete feature and thus, we hypothesize, represent popular queries of real-world interest. Such questions also trigger `answer boxes' in the search results, containing responses deemed best by Google, which we extract and refer to as Google answers.
Our human evaluation (\S\ref{subsec:data:quality}) found the collected questions and answers to be of high quality (over 94\% valid answers). 

\name\ provides a unified test bed to study inherent differences between questions. To do so, we fine-tune generative pre-trained language models (LMs)~\cite{lewis2020bart,raffel2020exploring} on different subsets of \name, and ask whether models trained for different answer types:
\begin{enumerate}[noitemsep, label=(\subscript{Q}{{\arabic*}}),leftmargin=35pt]
    \item \label{q1} \emph{benefit similarly from pre-training?}
    \item \label{q2} \emph{benefit similarly from labeled data?}
    \item \label{q3} \emph{benefit similarly from larger models?}
\end{enumerate}

To understand the contribution of pre-training, {\bf \ref{q1}}, we train the powerful T5 language model~\cite{raffel2020exploring} on \name{}   with a small amount of labeled data. While LMs struggle, as expected, in this setting on \emph{short} response questions, they perform surprisingly well in generating \emph{snippet} and \emph{collection} responses.\footnote{\changed{Over 30-40\% of our best model's snippet and collection answers were preferred by crowdworkers over Google's answers; achieving 50\% here would mean parity with Google}.}
We hypothesize this is because response fluency and coherence have a much higher weight in such questions, and these factors remarkably benefit from the LM pre-training objective. 
Regarding the value of labelled data, {\bf \ref{q2}}, we observe the opposite trend: \emph{short} response questions benefit consistently from increasing amounts of supervised (labeled) data, whereas both \emph{snippet} and \emph{collection} response questions show minimal gains (e.g., only 5-10\% improvement when going from 2k training examples to 200k or even 2 million).
Lastly, on the benefit of model size, {\bf \ref{q3}}, we find larger models to be more effective in all cases as expected, but the gains are much more pronounced for \emph{snippet} and \emph{collection} response generation (20+\%) as compared to \emph{short} responses (5-10\%), under human evaluation. 

Additionally, we expect \name{} to facilitate further research on models for answering snippet and collection response questions.  While the largest models we consider score surprisingly high on these questions, they
are still far from reaching Google's quality under either automated or human evaluations. Importantly, due to little benefit observed from more labeled data on such questions, further progress requires rethinking the approach and devising new solutions.

Lastly, we find \name{} to be a valuable resource for training models. On the long-answer dataset ELI5~\cite{fan2019eli5}, T5 trained only on our snippet questions performs on par with state-of-the-art models trained on ELI5 data.

\changed{
Our closing remarks describe why we aren't simply replicating an existing QA system at Google, place our findings in context, and discuss future uses of \name{}, such as creating a neural knowledge-base or a question generation system. 
}

\paragraph{Contributions.} Our contributions are threefold:
\begin{enumerate}[itemsep=1mm,noitemsep,leftmargin=.5cm]
    \item We present \name, a collection of 3 million question-answer pairs with a diverse set of answers, along with a crowdsourced assessment of its quality.
    \item We benchmark state-of-the-art models on \name{}, both in terms of automatic and human judgments, and observe remarkable differences in how models behave on different answer types.
    \item We demonstrate that \name{} is also a valuable model training resource by showing strong 
generalization to ELI5~\cite{fan2019eli5}. 
\end{enumerate}

\section{Related Work}
\label{sec:related:work}


A closely related work is the Natural-Questions (NQ) dataset \cite{kwiatkowski2019natural,lee2019latent} 
which contains questions written by Google users, and answers that were manually extracted from Wikipedia articles. 
While our questions (extracted via autocomplete) were also likely frequently asked by Google users, our dataset represents a different and wider distribution of questions (\S\ref{subsec:statistics}), likely because it encompasses different classes of answers, particularly snippet and collection responses.  Specifically, while NQ is dominated by `who', `when', and `how many' questions (cf.~Fig.~\ref{fig:bigram:distributions}(d)), \name{} has notably few `who' questions and a substantial portion of questions starting with `how to', `what is', `what does', `can you'. 


One notable QA dataset with long-form responses is ELI5~\cite{fan2019eli5,krishna2021hurdles}, containing questions/answers mined from Reddit forums. In contrast, \name{} is collected differently and is several orders of magnitude larger than ELI5. Empirically, we show that models trained on \name{} transfer surprisingly well to ELI5 (\S\ref{subsec:eli5}), indicating \name's broad coverage.

It is worth highlighting that there is precedent for using search engines to create resources for the analysis of AI systems. 
Search engines harness colossal amounts of click information to help them effectively map input queries to a massive collection of information available in their index~\cite{brin1998anatomy,joachims2002optimizing,berant2013semantic,joachims2017accurately}. 
Although academic researchers do not have direct access to information collected from the users of search engines, 
search results can act as a proxy for them and all the complex engineering behind them. In particular, the \name{} dataset used in this study probably is \emph{not} representative of \emph{a single} QA system in Google; on the contrary, we hypothesize, this data is produced by a complex combination of many systems, various forms of user feedback, as well as expert annotation/verification of highly popular responses. 

\section{\name{} dataset}
\label{sec:collection}

We describe how \name{} was collected, followed by dataset statistics and quality assessment.

\subsection{Dataset Construction}
\label{subsec:construction}

Constructing this dataset involved two main steps, extracting questions from search auto-complete and extracting answers from answer boxes.

\subsubsection{Query Extraction}
\label{subsubsec:query:extraction}

To extract a rich yet natural set of questions we use Google auto-completion.\footnote{http://google.com/complete/search?client=chrome\&q=...}
A similar strategy was also used by~\citet{berant2013semantic}, albeit in the context of a slightly different study. 
We start with a seed set of question terms (e.g., `who', `where', etc.; the complete list is in Appendix~\ref{appendix:query:terms}.) 
We bootstrap based on this set, by repeatedly querying prefixes
of previously extracted questions, in order to discover longer and richer sets of questions.
Such questions extracted from the autocomplete algorithm 
reflect popular questions posed by users of Google. 
We filter out any questions shorter than 5 tokens as they are often incomplete questions. 
This process yields over $\sim$5M questions, which were collected over a span of 6 months. 
The average length of the questions is about 8 tokens.  

\subsubsection{Answer Extraction}
\label{subsubsec:answer:extraction}
To mine answers to our collected questions, we extract the Google answer boxes shown on top of the search results when the questions are issued to Google. 
There are a variety of answer boxes. 
The most common kind involves highlighted sentences (extracted from various websites) that contain the answer to a given question. 
These form the \emph{snippet} and \emph{collection} answers in \name{}. 
In some cases, the answer box shows the answer directly, possibly in addition to the textual snippet. 
Similarly, \emph{unit-conversion} and \emph{time-conversion} they each have distinct answer boxes. 
Some technical details of the answer extraction is included in Appendix~\ref{sec:answer:extraction}. 

After the answer extraction step, we have all the necessary information to create a question in \name{}, such as the examples in Fig.~\ref{fig:examples}.

\paragraph{Answer Type Categories.}
    We use the HTML tags of the search results to infer answer type tags for each answer. 
    The overall list of types are shown in Table~\ref{tab:statistics} (examples in Fig.~\ref{fig:examples}). 
    We define `short' response questions to be the union of `knowledge', `unit-conversion', `time-conversion', and short answers from the `snippet` responses. 

Table~\ref{tab:statistics} summarizes various statistics about \name{} broken down into different question/answer types. 
Of the 5M collected questions, about half resulted in successful answer extraction from answer boxes. 
The largest type of questions received `snippet' 
answers with over 2.7M responses (examples shown in the left-most column of Fig.~\ref{fig:examples}).
The other major category is `collection' answers with 329k questions (examples shown on the right-most column of Fig.~\ref{fig:examples}).

\begin{table}[ht]
    \centering 
    \small 
    \begin{tabular}{lccc}
        \toprule 
        Answer types & Count & \makecell{\% of valid \\questions}  & \makecell{\% of valid \\answers}   \\
          \cmidrule(lr){1-1} \cmidrule(lr){2-2} \cmidrule(lr){3-3} \cmidrule(lr){4-4}
        
            short answers & 275k &-  & -  \\
            \ \ \rotatebox[origin=c]{180}{$\Lsh$} unit conversion  & 45k & 96.9 & 90.9 \\
            \ \ \rotatebox[origin=c]{180}{$\Lsh$} time conversion  & 2.5k & 93.2 & 70.9  \\ 
            \ \ \rotatebox[origin=c]{180}{$\Lsh$} knowledge  & 32k & 96.3 & 84.1 \\ 
            \ \ \rotatebox[origin=c]{180}{$\Lsh$} snippet (short) & 196k & 98.4 & 76.0  \\
            snippet  & 2.7M  & 98.5 & 95.5 \\ 
            collection & 329k & 99.7 & 98.9 \\ 
            \midrule
            Overall &  3.1M & 98.6 & 94.5 \\ 
        \bottomrule 
    \end{tabular}
    \caption{
    Statistics of different answer types in \name{} (\S\ref{subsec:statistics}) and their quality evaluation by crowdworkers (\S\ref{subsec:data:quality}). 
        According to human ratings, a very small percentage of the questions are invalid (first column). Among the valid questions, a substantial portion are deemed to have valid answers. 
    }
    \label{tab:statistics}
\end{table}

\subsection{Quality Assessment of \name}
\label{subsec:data:quality}
We perform a crowdsourcing experiment to assess the quality of the extracted questions and their answers. 
We use Amazon Mechanical Turk (AMT) to annotate about 2.5k randomly selected question-answer pairs. 
The annotators were asked to annotate (1) whether a given question makes sense and, if so, (2) whether the provided answer is 
complete. 

\paragraph{Annotation details.}
Since our task is focused on English, we required workers to be based in a country with a population predominantly of native English speakers (e.g., USA, Canada,
UK, and Australia) and have completed at least 5000 HITs with
$\geq$ 99\% assignment approval rate. Additionally, we have a qualification test with half-a-dozen questions all of which need to be answered correctly by our annotators. 
To prevent biased judgements, we also ask annotators to avoid using Google search (which is what we used to  mine \name) when annotating the quality of shown instances.
Each example is annotated by 3 independent annotators and aggregated via a majority vote of the 3 labels.

\paragraph{Assessment results.}
We compute aggregate statistics for (1) average rating of questions and (2) average rating of the answer quality, among valid questions. 
As can be seen in the results in Table~\ref{tab:statistics} only a small percentage of the questions were deemed `invalid'. Additionally, among the `valid' questions, a high percentage of the answers were deemed high-quality for most of the question/answer types. 
This indicates a reasonable quality of \name{} question-answer pairs, as evaluated directly,
independent
from any systems. 
(Examples of invalid questions/answers are provided in Appendix~\ref{appendix:invalids}.) 

\subsection{Dataset Analysis}
\label{subsec:statistics}

\begin{figure}[ht]
    \centering
    \begin{subfigure}[b]{0.23\textwidth}
        \caption{\name{}}
        \includegraphics[scale=0.5]{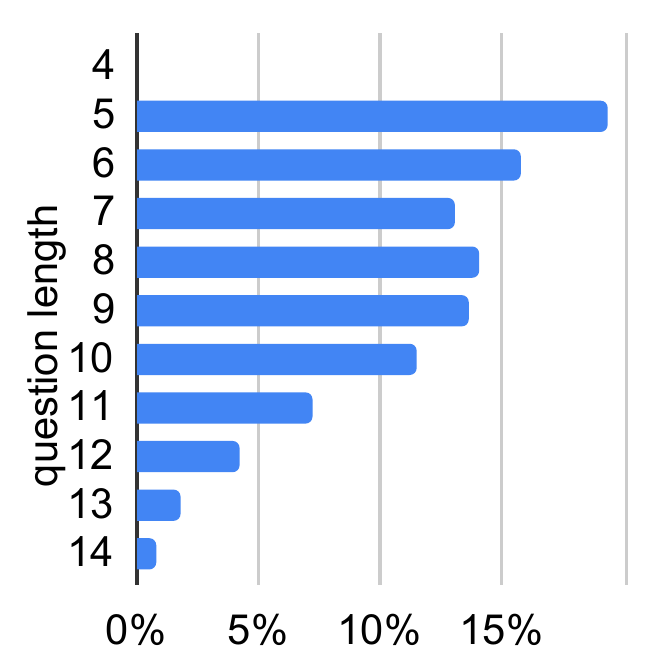}
    \end{subfigure}
    \begin{subfigure}[b]{0.23\textwidth}
        \caption{ 
            NQ
        }
        \includegraphics[scale=0.5]{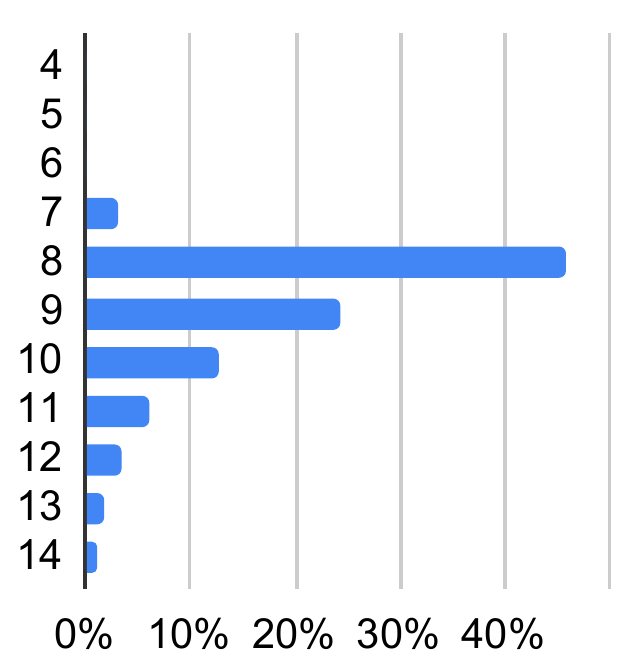}
    \end{subfigure}
    \caption{Comparison of question length distributions}
    \label{fig:len:distt}
\end{figure}

To better understand the content of \name, we present several distributions from the data. 
Fig.~\ref{fig:len:distt} shows the length distribution of \name{} questions and that of 
NQ~\cite{kwiatkowski2019natural}. 
While a vast majority of NQ questions contain 8-10 tokens, \name{} questions have a somewhat broader range of lengths.

\begin{figure*}[t]
    \centering
    \captionsetup[subfigure]{justification=centering}
    \begin{subfigure}[b]{0.22\textwidth}
        \caption{\name{}\\\emph{Short} answer questions}
        \includegraphics[scale=0.52]{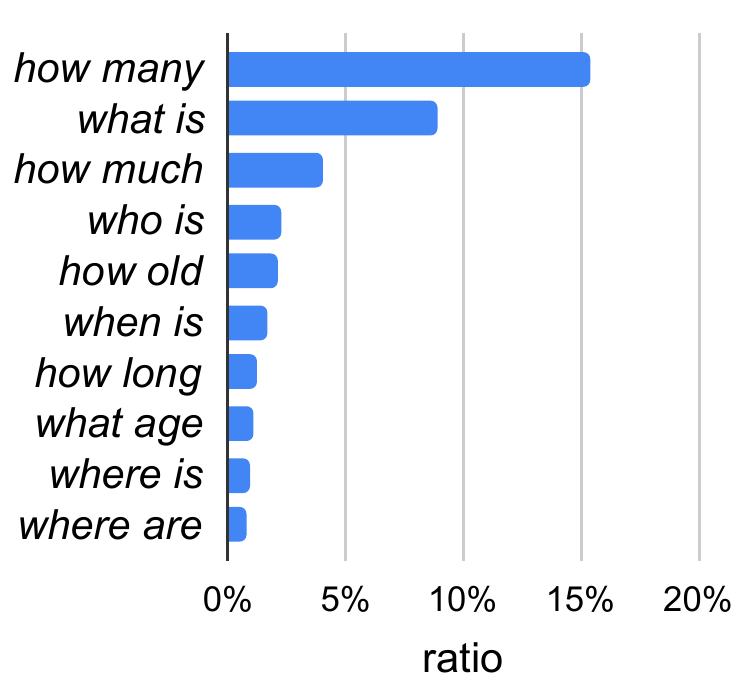}
    \end{subfigure}
    \hspace{1ex}
    \begin{subfigure}[b]{0.22\textwidth}
        \caption{\name{}\\\emph{Snippet} questions}
        \includegraphics[scale=0.52]{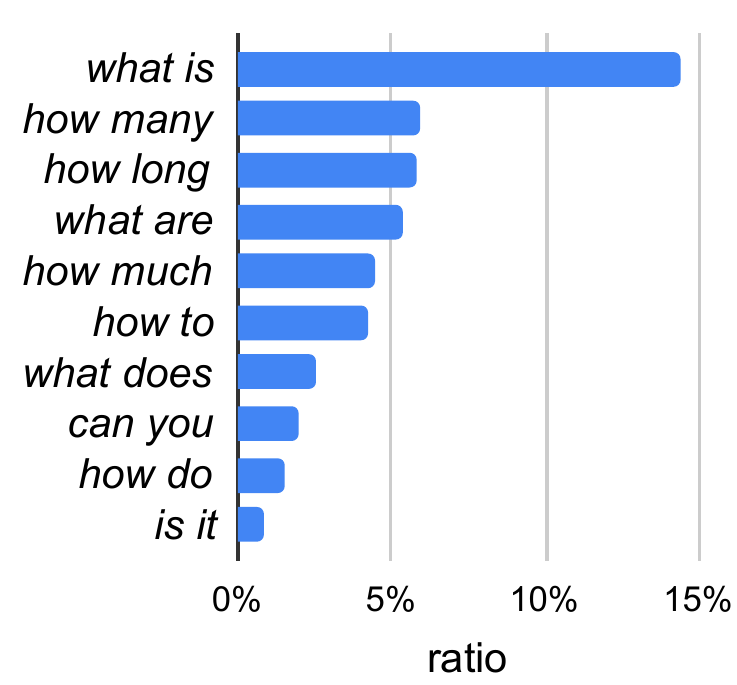}
    \end{subfigure}
    \hspace{1ex}
    \begin{subfigure}[b]{0.23\textwidth}
        \caption{\name{}\\\emph{Collection} questions}
        \includegraphics[scale=0.52]{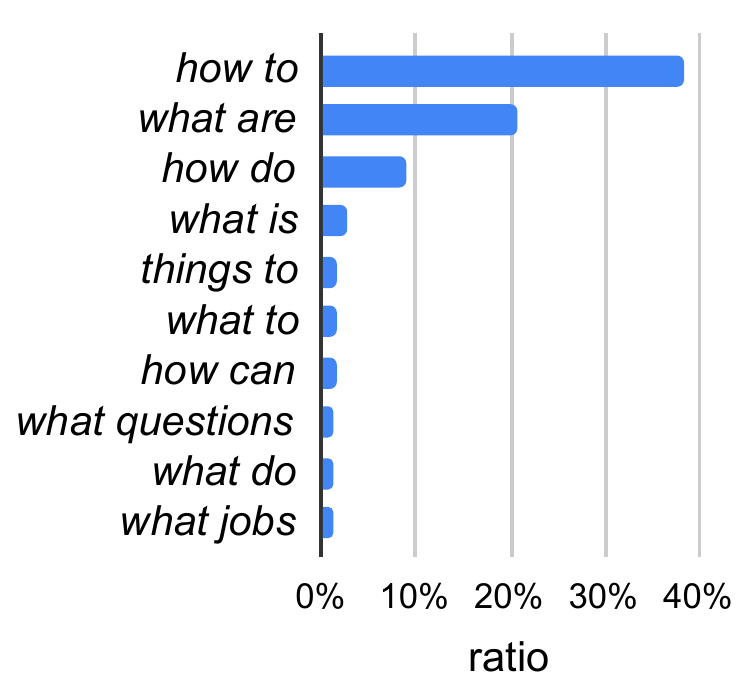}
    \end{subfigure}
    \hspace*{5ex}  
    \begin{subfigure}[b]{0.22\textwidth}
        \caption{
            NQ
        }
        \includegraphics[scale=0.52]{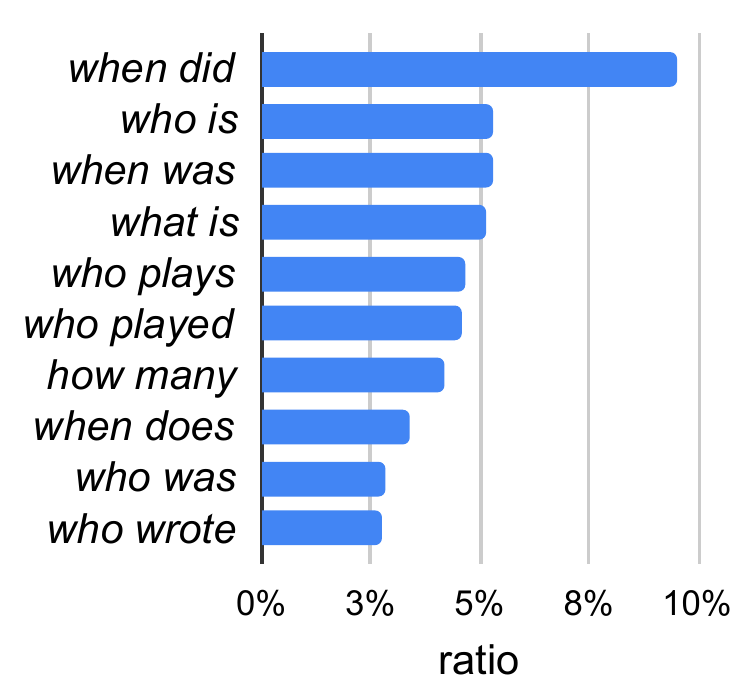}
    \end{subfigure}
    \caption{The distribution of common bigrams in questions of \name{} (a,b,c) vs.\ 
    NQ
    (d).
    }
    \label{fig:bigram:distributions}
\end{figure*}

To gain insights about the type of questions, we study the distribution of the most frequent opening bigrams of the questions (Fig.~\ref{fig:bigram:distributions}).
Among the \emph{short} answer questions, the majority are information-seeking questions about counts (`how many'), places (`where is'), values (`how much'), and people (`who is'). 
They also include `what is' questions, which can cover a wide variety of open-ended queries with short answers (e.g., \emph{what is the time difference $\dots$?}, \emph{what is the length of X?}, etc.).
Among the \emph{snippet} questions, the dominant pattern is `what is', which typically is an open-ended question about entities (e.g., \emph{`what is X?'} or \emph{`what is the difference between X and Y?'}). Among the \emph{collection} response questions, most questions are about steps or ingredients needed to accomplish a goal (`how to' and `what are'). A comparison with the bigram distribution of NQ (Fig.~\ref{fig:bigram:distributions}; right) highlights that \name{} represents a different and wider class of questions.
Specifically, NQ has many `who', `when', and `how many' questions, while \name{} dominantly contains `how' and `what' questions, which typically require explanatory responses. 

\changedD{
In terms of the different reasoning types, \name{} has an extremely long-tail of reasoning challenges, due to our data collection procedure. For example, we observed many challenges such as application of mathematical definitions (Q: \emph{`what is the multiplicative inverse of 10?'} A: \emph{`1/10'}), linguistic definitions (Q: \emph{`a man who looks after cattle?'} A: \emph{`cowherd'}; Q: \emph{`a man who protects sheep?'} A: \emph{`Shepherd'}), comparisons (Q: \emph{`are boiling and evaporation the same?'}; Q: \emph{`what is the difference between night sky and day sky?'}), instantiation (Q: \emph{`what is an example of kinetic energy?'}), etc., to name a few. 
Because of the long tail of reasoning phenomena, a detailed analysis would require careful human annotations which we leave for future work. 
}

\definecolor{electricpurple}{rgb}{0.75, 0.0, 1.0}
\definecolor{azure(colorwheel)}{rgb}{0.0, 0.5, 1.0}

\begin{figure*}[th]
    \centering
    \begin{subfigure}[b]{0.32\textwidth}
        \caption{\task{short} task}
        \includegraphics[scale=0.53]{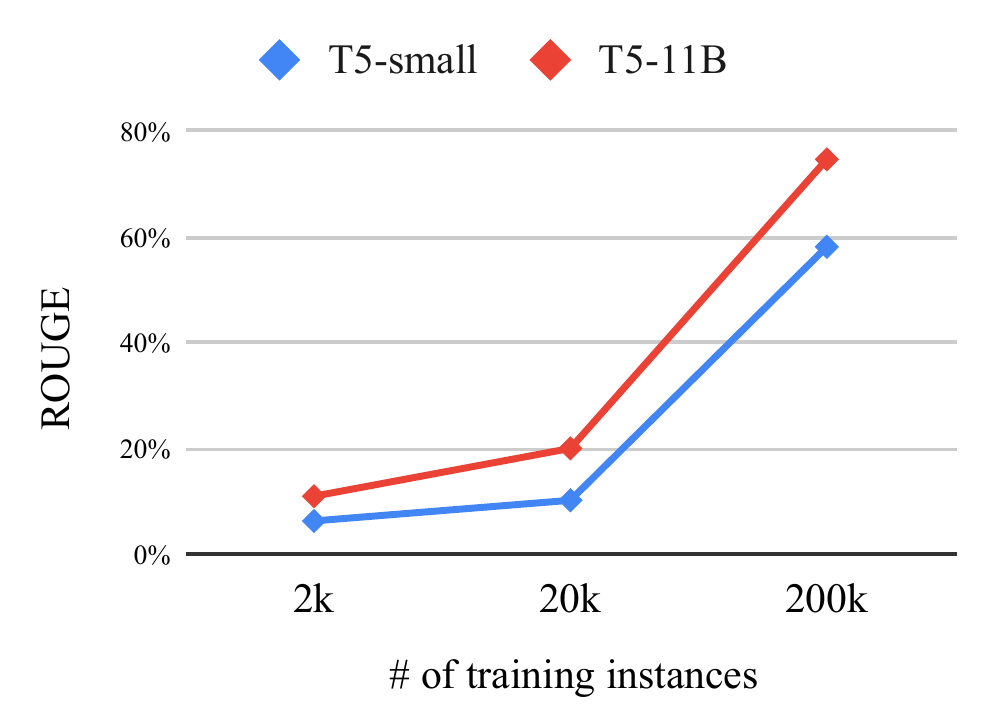}
    \end{subfigure}
    \begin{subfigure}[b]{0.32\textwidth}
        \caption{\task{snippet} task}
        \includegraphics[scale=0.53]{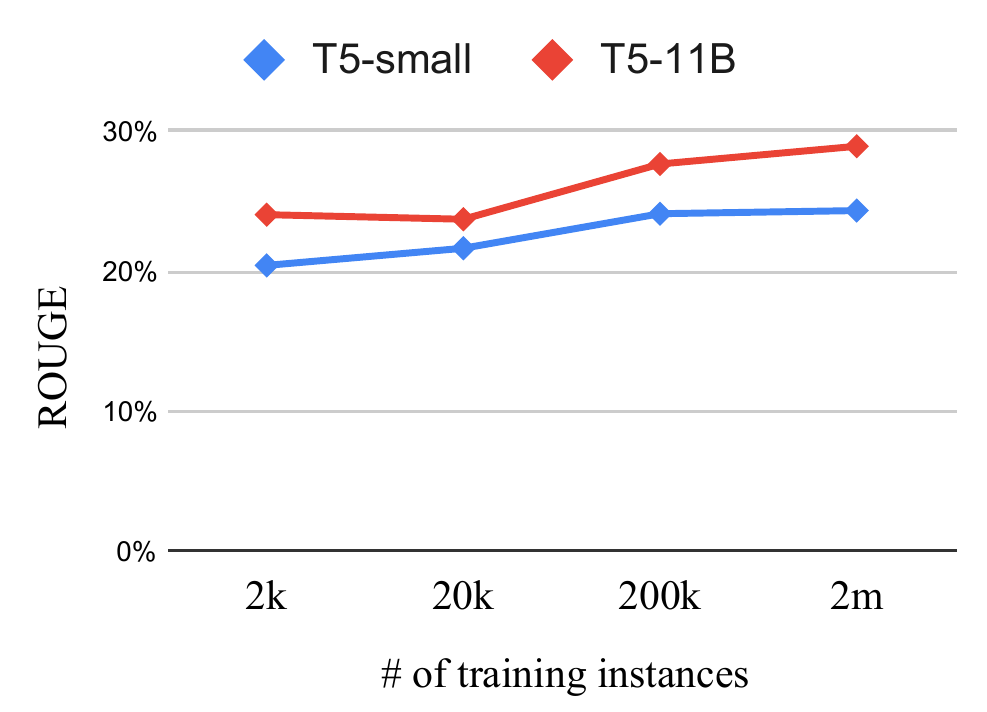}
    \end{subfigure}
    \begin{subfigure}[b]{0.32\textwidth}
        \caption{\task{collection} task}
        \includegraphics[scale=0.53]{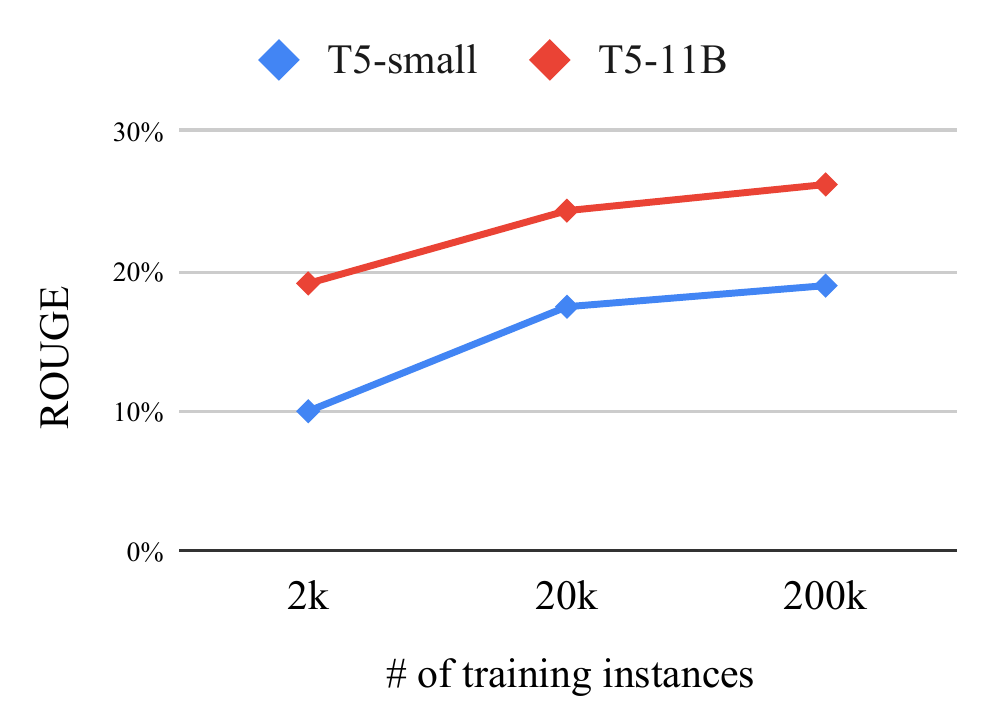}
    \end{subfigure}
    \begin{subfigure}[b]{0.32\textwidth}
        \includegraphics[scale=0.245,trim=0.3cm 5.5cm 0cm 6.0cm,clip=true]{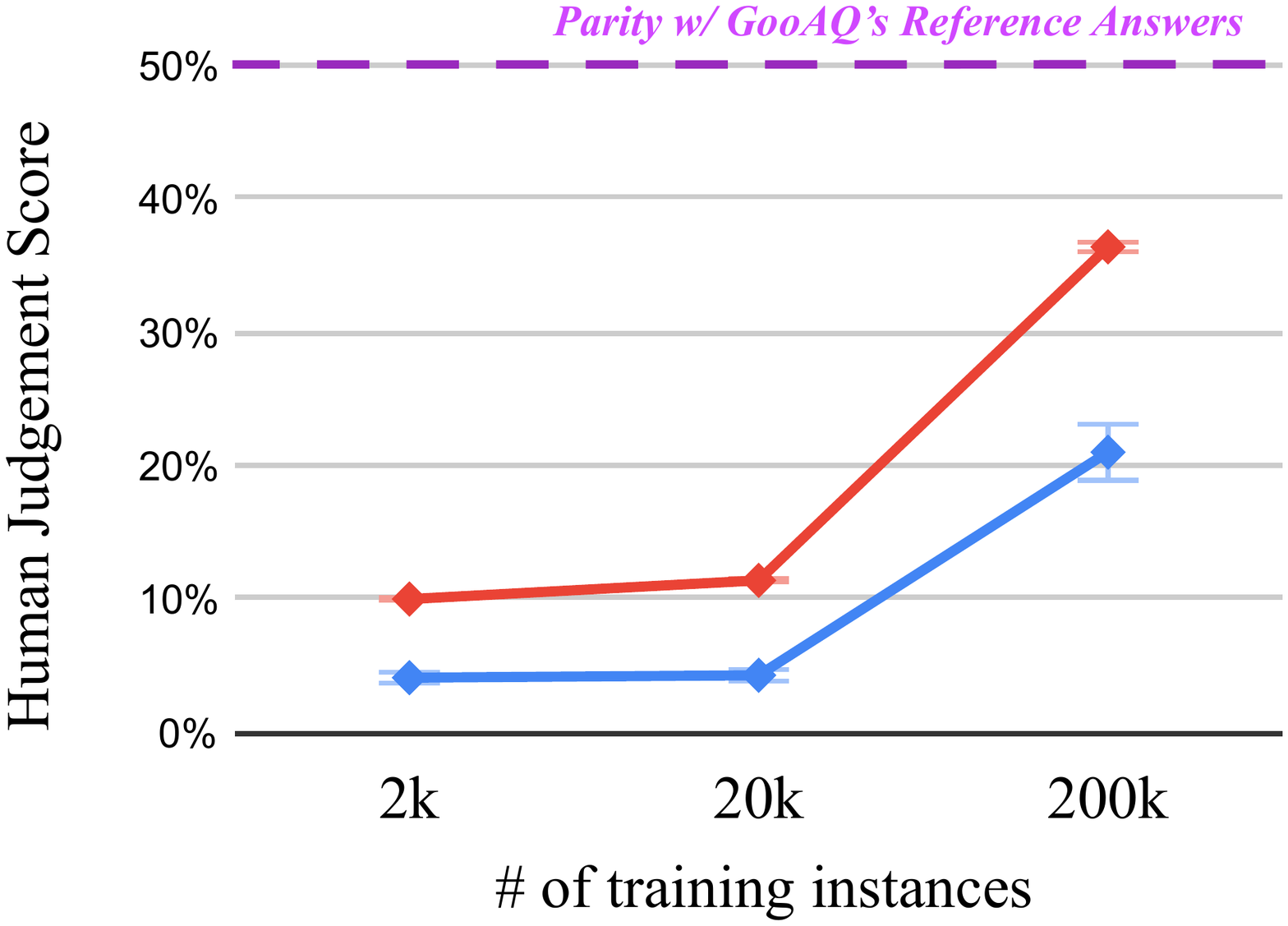}
    \end{subfigure}
    \begin{subfigure}[b]{0.32\textwidth}
        \includegraphics[scale=0.245,trim=0.3cm 5.5cm 0cm 6.0cm,clip=true]{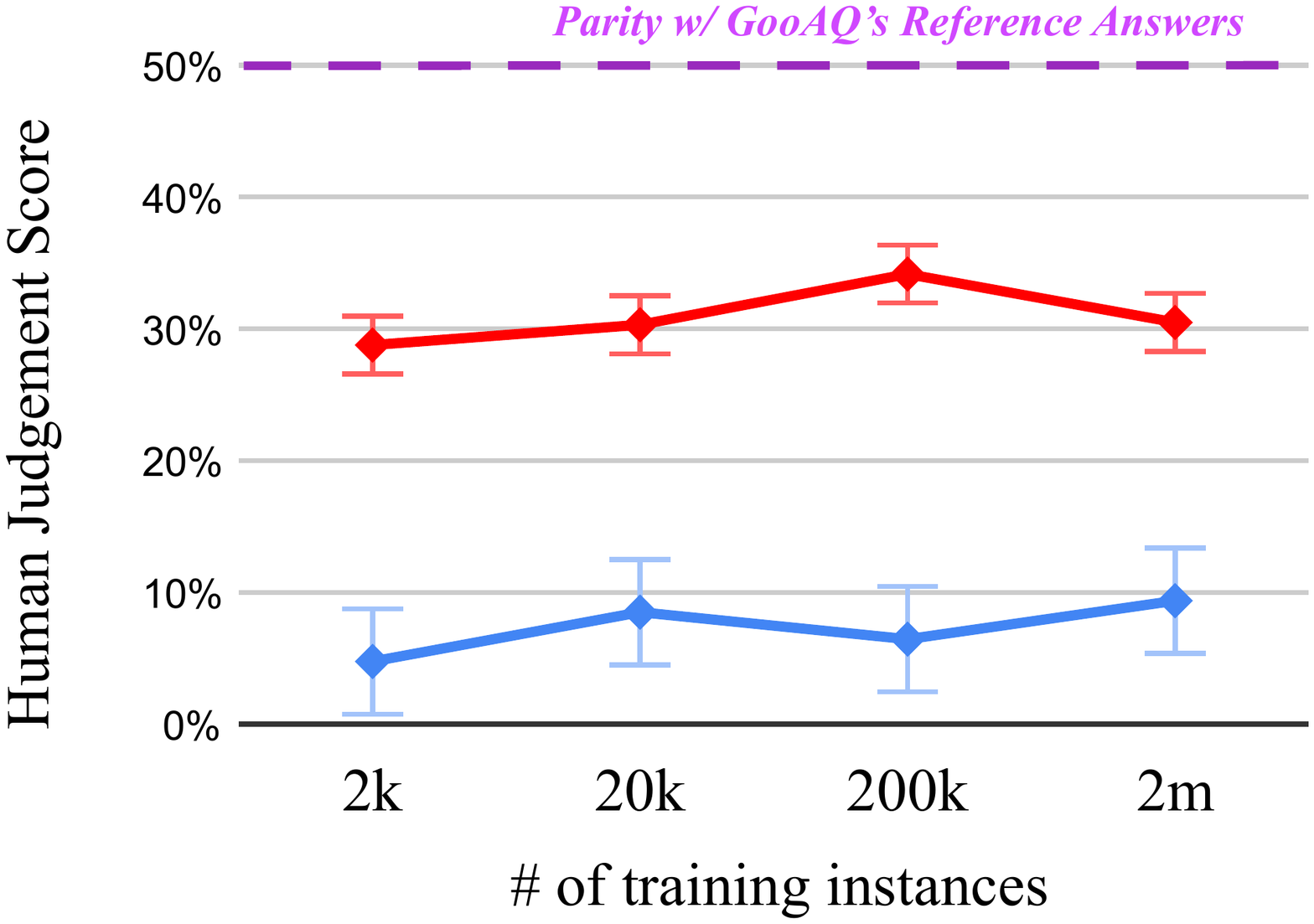}
    \end{subfigure}
    \begin{subfigure}[b]{0.32\textwidth}
        \includegraphics[scale=0.245,trim=0.3cm 5.5cm 0cm 6.0cm,clip=true]{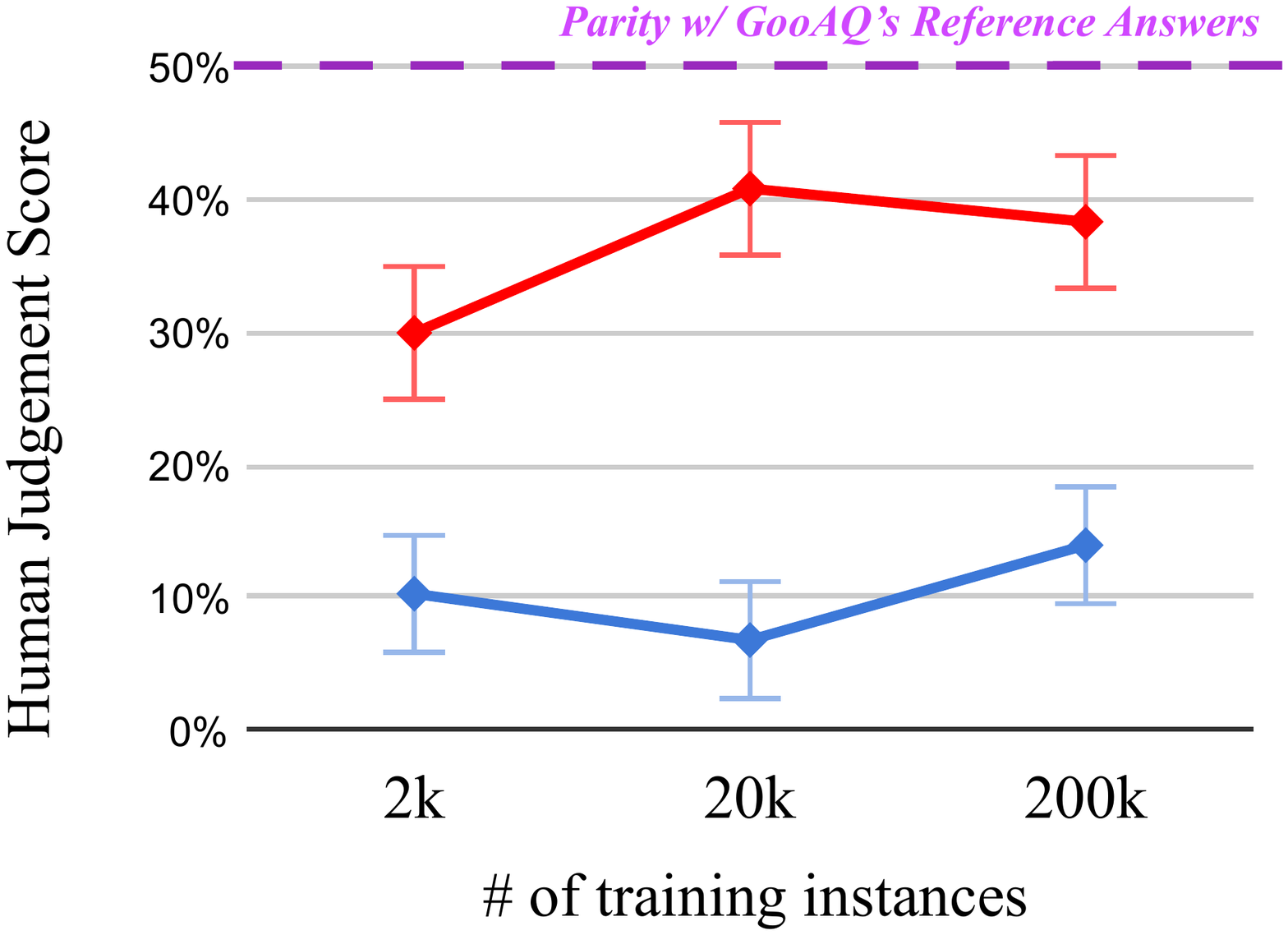}
    \end{subfigure}
    \caption{
        Evaluation of T5 ({\color{azure(colorwheel)}small},{\color{red}11B}) models on different sub-tasks of \name{} via \emph{automatic} metrics (top) and  \emph{human} judgements (bottom). 
        For human evaluation, 50\% is the border at which the model output and the ground truth responses are {\color{electricpurple}indistinguishable}. 
        The short-answer sub-tasks (\task{short}; left) have a relatively low performance when supervised with $2k$ instances. However, they benefit more than the long-answer sub-tasks (\task{snippet} \& \task{collection}) from more labeled data. 
        Additionally, we observe that the gap between the two systems is bigger in terms of human evaluation (compared to the corresponding gap in terms of automatic evaluation), especially in the \emph{long} response tasks (middle \& right). 
    }
    \label{fig:eval}
\end{figure*}


\begin{figure}[htb]
    \centering
    \egbox{
    \includegraphics[scale=0.57,trim=-0.45cm 0.1cm 0.2cm 0.29cm]{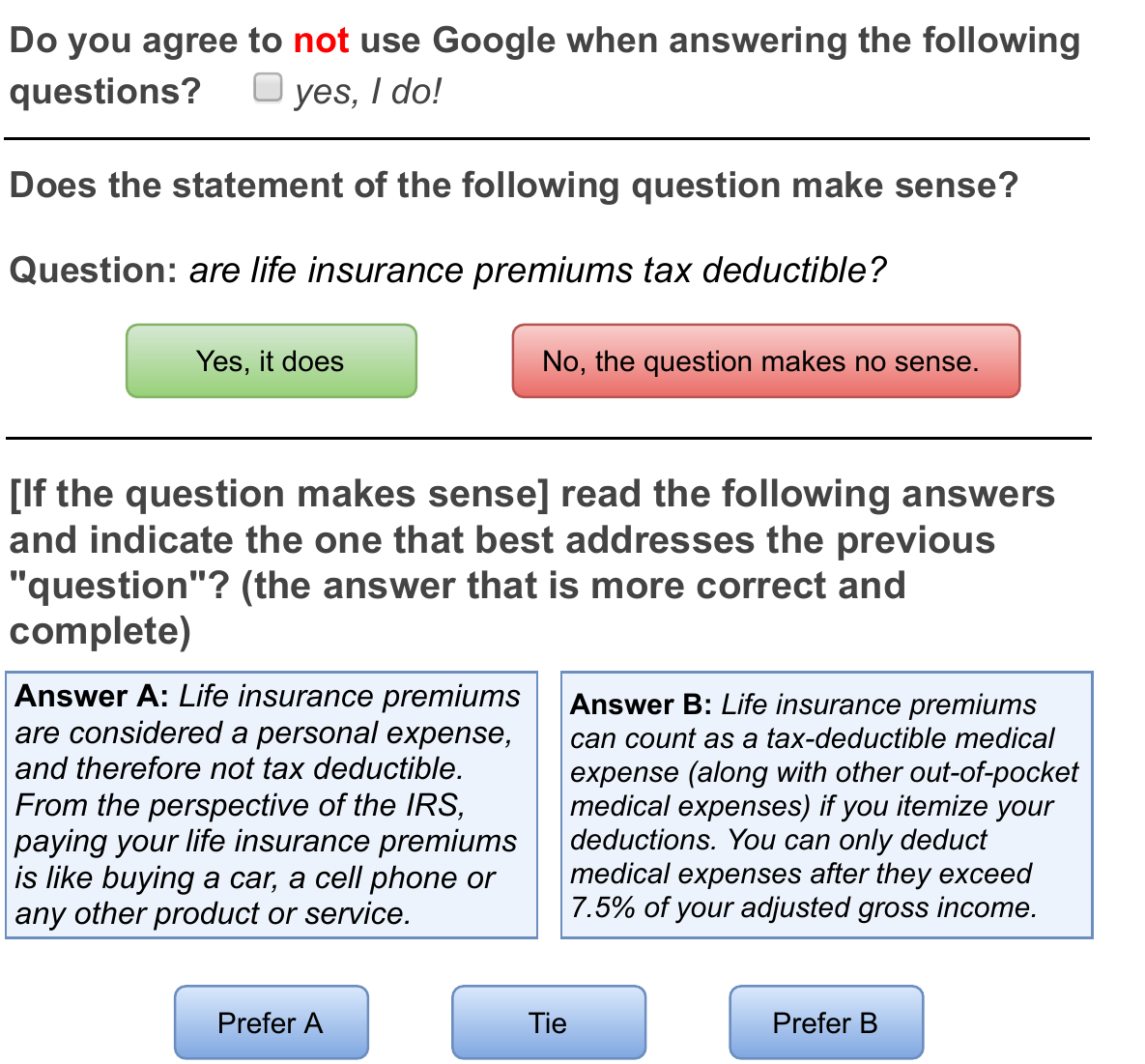}
    } \vspace{-.3cm}
    \caption{Crowdsourcing interface used for  human assessment of our baselines (\S\ref{sec:experiments}). 
    We use a similar template (with a single answer) to estimate the quality of \name{} (\S\ref{sec:collection}). 
    }
    \label{fig:crowdsourcing:picture}
\end{figure}

\section{Task Setup and Models}
\label{sec:experiments} 

\name{} naturally forms a dataset for the task of open QA, where the input is a question and the output is its answer. 
Unlike the reading comprehension setting, the context for answering the question is not provided as part of the input. 
In particular, we consider the so-called 
`closed-book' setup~\cite{roberts2020much} where the model (e.g., a language model) is expected to use background knowledge stored within it, without access to any additional explicit information retrieval mechanism.\footnote{
\changedD{
In our early experiments, we considered information-retrieval (IR) systems in conjunction to LMs (i.e., an `open-book` setup). We observed that IR results are quite noisy for most open questions. Hence, a system trained with the retrieved documents did not benefit from them (the model learned to ignore the noisy retrieval results). Similar observations were also made by \citet[Sec3.1]{krishna2021hurdles} (``generations are similar irrespective of type of retrievals'').
}
}

\subsection{Problem Setup}
We split \name{} into three sub-tasks: (\task{short})  \emph{short} responses questions,  
(\task{snippet})  \emph{snippet} responses questions,  and (\task{collection}) \emph{collection} response questions. 
We train and evaluate models for each of these sub-tasks separately. We define them as different sub-tasks since by merely reading the questions it might not be clear whether its response should be short, a snippet, or a collection,

\paragraph{Data splits.}
For each sub-task, we randomly sample \emph{test} and \emph{dev} sets such that each evaluation split contains at least 500 instances of each response type. 
We experiment with varying training data sizes to better understand the value of labeled data. 
\citet{lewis2020question} have shown that leakage from training data to the evaluation sets often results in unrealistically high scores. 
To minimize this issue, we create training splits by selecting the most \emph{dissimilar} instances to our evaluation splits. 
The measure of \emph{similarity} for each training instance is computed as the maximum amount of token-overlap with any of the instances in the test/dev set (computed over both questions and answers). 
Using the most \emph{dissimilar} subset of the training instances, we create training splits of the following sizes: 2k, 20k, 200k. 
For \task{snippet}, we also have a 2M training set since this sub-task has more data.

\subsection{Evaluation Metrics}
\label{subsec:evaluation}
\paragraph{Automatic evaluation.}
We use the ROUGE-L metric~\cite{lin2004rouge} 
, which is a common metric
for assessing the quality of models for text generation tasks. 
The results of the automatic evaluation for each sub-task are shown in the top row of Fig.~\ref{fig:eval}. 

\paragraph{Human evaluation.} We additionally perform human evaluation which is generally known to provide more accurate evaluation for generated text. Specifically, we ask  crowdworkers to indicate if they prefer the predicted answer by the model or the Google answer for each question (without revealing the source of the answers).

The annotation interface is shown in Fig.~\ref{fig:crowdsourcing:picture}, which 
is essentially the same template used for the quality assessment of the dataset (\S\ref{subsec:data:quality}), except that here the crowdworkers are shown a \emph{pair} of responses for each question---the reference answer (extracted from Google) and the one generated by the model---turning the task into a \emph{comparative} one. Before annotating each instance, we remind the annotators to avoid using Google. 
Then we ask them to check if the provided question is clear enough and makes sense. 
Upon indicating `yes', they choose between the Google answer, the generated answer by our model, or indicate that they are equally good (by selecting `tie').  

For each question, we obtain annotations from 5 independent annotators and aggregate via a majority vote.\footnote{Ties occurred infrequently (e.g., in 6\% of the cases when evaluating our largest T5 model) and were broken at random.}
The model receives a credit of 1 if the majority vote favors the model's prediction, 0.5 if the majority vote is the `tie' label, and 0 otherwise.
The overall accuracy score for the model is computed by averaging instance-level scores, after discarding questions annotated as invalid (`this question makes no sense').

The resulting  \emph{human-evaluation} metric indicates how often were model predictions preferred over Google's answers.  
In this evaluation, 50\% is the mark where the annotators are not able to distinguish the model's responses from Google's answers in any meaningful way. 
The results of human evaluation are shown in the bottom row of Fig.~\ref{fig:eval}.


\subsection{Models}
For our evaluation, we use the T5 model~\cite{raffel2020exploring}, a recent text-to-text framework that has achieved state-of-the-art results on a variety of tasks, including open QA~\cite{roberts2020much}. 
The models are trained to produce answer string, given the question string as input. 
We use two model sizes that capture the two extremes: the smallest model (`small') and the largest model (`11B').
Both models were trained for 20k steps on the training splits, dumping checkpoints every 2k steps (with 196,608 tokens per batch on v3-128 TPUs) with the default hyperparameters.
We select the checkpoint with the highest score on the `dev' set and report its corresponding `test' score.

\section{Empirical Results and Analyses}
In this section, we evaluate the behavior of models for various answer types (\S\ref{subsec:behavior:various:types}). We further show how \name{} can support research in answering questions with long answers (\S\ref{subsec:challenge};\S\ref{subsec:eli5}).

\subsection{Models vs. Various Answer Types}
\label{subsec:behavior:various:types}

\paragraph{\ref{q1} Model pre-training is surprisingly effective on the snippet and collection answer sub-tasks}
Both automatic and human evaluations of these two classes of questions (Fig.~\ref{fig:eval}; middle \& right) demonstrate that the T5-11B model is surprisingly effective at answering them, 
with only 2k training examples.
For example, crowdworkers even prefer the model's answer over Google's in 30\% of the cases.\footnote{Across all experiments, the model's and Google's answers were deemed a ``tie'' in fewer than 10\% of the cases.}
This is in contrast with short answer questions, where the model's accuracy is only around 10\% and crowdworkers prefer Google's answers in about 90\% of the cases.



To understand this observation, one needs to put into perspective several factors that are at play. First, short answer questions typically ask for encyclopedic knowledge and, therefore, \emph{correctness} of the answers matters the most. In snippet and collection questions, we suspect \emph{coherence} of the response carries a heavier weight. 
This is partly due to the nature of the questions, which can be responded to in a variety of ways.
For example, the snippet response to the question of \emph{how many calories burned 30 minutes crossfit?} (Fig.~\ref{fig:examples}) could refer to a range of calorie consumption, depend on the choice of activity during crossfit, or vary by the attributes of the person working out. All of these responses would be equally correct.

\paragraph{\ref{q2} Labeled data is more helpful for short answer questions.}
Based again on both the automatic and human evaluations (Fig.~\ref{fig:eval}; left), the performance of both small and 11B parameter models on the short response questions quickly improves as we increase the amount of training data, especially beyond 20k. This is in contrast with snippet and collection questions, where even 200k labeled instances don't appear to help much, indicating that in these question types, model pre-training contributes more than labeled data does.

\paragraph{\ref{q3} Human evaluation accentuates the gap between the `small' and `11B' models, especially on snippet and collection response questions.} This is visually evident from the gap between the blue and red curves in the bottom row vs.\ the top row of Fig.~\ref{fig:eval}.
This is compatible with recent work of \citet{min2021neurips}, who also observed that the gap between two reasonably different systems is bigger when using human evaluation. 
We hypothesize this is due to the crudeness of automatic evaluation metrics, and an indication of the necessity of human evaluation to distinguish between nuanced differences among generated responses.

What is perhaps more interesting (and not evident from prior work) is that the gap between automatic and human evaluation is larger for the snippet and collection questions than short answer questions, especially for the T5-small model. This is, at least partly, due to the inaccuracy of automatic metrics in evaluating long text.

\subsection{\name{} as a challenge for LMs}
\label{subsec:challenge}
One can view \name{} as a challenge for NLP, for building self-contained models that achieve performance comparable to Google's answers. 

As mentioned earlier, our human evaluation measures the comparative quality of the model predictions and our reference responses (Google's answers). 
Hence, a value of 50\% in this evaluation is an indication that the predictions are on par with (i.e., indistinguishable from) the ground-truth responses (defined in `human-evaluation' \S\ref{subsec:evaluation}). 

As the bottom row of Fig.~\ref{fig:eval} shows, the T5-11B model comes quite close to Google's answers but is still not quite at par with it. We hope this gap will encourage further research in building stronger models, especially for the snippet and collection answer questions where more labeled data doesn't appear to be a promising way to increase accuracy.

\subsubsection{Error Analysis}
\label{subsec:error:analyais}

To gain an intuition about the mistakes made by the models, we conducted a small-scale errors analysis of model predictions. For each model, we (one of the authors) annotated 30 predictions, and labeled them with the following error categories inspired from existing evaluations of text summarization~\cite{chaganty2018price}: \emph{incompleteness}, indicating the lack of expected substance in the prediction; \emph{redundancy}, indicating repeated content; \emph{hallucination}, indicating existence of made-up statements; and \emph{incoherence} indicating the existence of grammatical errors (examples in Appendix~\ref{subsec:error:analyais-appendix}).


\begin{table}[ht]
    \centering
    \small
    \begin{tabular}{ccccc}
         \toprule 
         Model & \rot{Incompleteness} & \rot{Redundancy} & \rot{Hallucination} & \rot{Incoherence}  \\
         \cmidrule(lr){1-1} \cmidrule(lr){2-2}   \cmidrule(lr){3-3}   \cmidrule(lr){4-4}   \cmidrule(lr){5-5} 
          T5-small &  52.5 & 65.0 & 47.5 & 2.5 \\ 
          T5-11B &  22.5 & 8.3 & 18.3 & 0.0 \\ 
          \bottomrule
    \end{tabular}
    \caption{Error distribution for the two models
    }
    \label{table:errors}
\end{table}

The results of our error analysis are summarized in Table~\ref{table:errors}. 
As expected, the `small' model makes more errors across all categories, and suffers particularly from \emph{redundancy} and \emph{incompleteness}. Overall, both models have very little \emph{incoherence}, which is to be expected from their strong pre-training.

\subsection{Extrinsic Utility of \name{}}
\label{subsec:eli5}

To showcase the value of \name{} as a model training resource,
we train our models on questions from \name{} and evaluate them  on ELI5~\cite{fan2019eli5}, a relatively recent dataset with long-answer questions extracted from Reddit posts.


\begin{table}[ht]
    \centering
    \small
    \resizebox{\linewidth}{!}{
    \begin{tabular}{lccc}
        \toprule
         Model & \makecell{Supervision} & \makecell{Uses\\IR?} & Score  \\
         \cmidrule(lr){1-1} \cmidrule(lr){2-2}  \cmidrule(lr){3-3}  \cmidrule(lr){4-4 } 
         T5-small & \name{} (no ELI5) & no & 21.7 \\ 
         T5-11B & \name{} (no ELI5) & no & 22.9 \\ 
         \midrule
         T5-small & ELI5 & no & 19.0 \\ 
         T5-11B & ELI5 & no & 22.7 \\ 
         \midrule 
         RAG$^*$ & ELI5 & yes & 14.1\\ 
         RT+REALM$^*$ & ELI5 & yes & 23.4\\ 
         \bottomrule
    \end{tabular}
    }
    \caption{
        Evaluation of our models on ELI5 dataset. 
        Results indicated with * are reported from prior work~\cite{krishna2021hurdles}. 
        T5 fine-tuned on \name{} performs well on ELI5, another long-answer dataset. 
    }
    \label{tab:eli5}
\end{table}

Our evaluation, summarized in Table~\ref{tab:eli5}, shows that both our small and 11B T5 models trained on \name{}'s snippet-answer subset
(no training on ELI5)
perform quite well (21.8 and 22.9, respectively) when evaluated on ELI5. 
They are even \emph{better than} the same architectures trained with ELI5's own training data (19.0 and 22.7, resp.) and on par with retrieval based state-of-the-art models (23.4).
    Complementary to these results, a T5-11B model trained on ELI5 and evaluated on \name{} results in 22.6\%, much lower than $\sim$28.9\% in Table~\ref{fig:eval}.

We hypothesize that despite \name{} being collected differently than ELI5, a notable portion of ELI5 is covered by \name, indicating good coverage of common questions posed by ordinary users.

\section{Closing Remarks}
\label{sec:closing-remarks}

We studied open QA under diverse response types. 
To this end, we collected \name{}, a very large set of QA pairs mined from Google, with a variety of short and long answer types, all of which are collected using a unified, coherent process, enabling a cross-type comparison. The auto-complete system used for our question collection likely reflects a natural distribution of questions asked by users. 

We benchmarked two variants of a state-of-the-art self-contained text generation model (T5, without retrieval) on three different sub-tasks of \name{}: short, snippet, and collection response questions. Our analysis, using both automatic and human evaluations, brings out the distinct behavior of LMs on long and short response questions. For example, while short response models benefit heavily from more labeled data, the surprisingly strong performance of long response models is driven mostly by their pre-training. We also demonstrate that \name{} is a valuable resource for training  models by showing high performance on an extrinsic task, ELI5, while using only \name{} data for training.

\paragraph{Scope of our conclusions.}
One must be careful in taking our specific conclusions out of the context of this study (i.e., the dataset at hand, the models, the evaluation metrics used, etc.). 
While we expect our findings to be fairly general, it may be possible to come up with a different long-form QA dataset where the trends across answer types differ.

Knowledge leakage across train and evaluation sets has been shown to significantly inflate performance numbers on recent open QA datasets~\cite{lewis2020question,emami2020analysis}. 
Similar concerns have motivated our careful training/evaluation splits of the data (\S\ref{sec:experiments}) and experiments with varying training set sizes. 
Nevertheless, we found it challenging to define (and assess) the amount of such leakage, and welcome such studies on \name{}.

\paragraph{Are we mimicking Google's QA?}
A reader might question the value of this work by noting that the website from which \name{} was mined had likely also used a QA system to begin with.
In other words, are we basically reverse-engineer Google's internal QA system~\cite{kilgarriff2007googleology}?

While we (the authors) are not aware of how Google answer box system works, we suspect that it is much more complex than a single QA system built using a single LM like T5~\cite{raffel2020exploring}. 
The system, besides incorporating one or more QA models, likely makes heavy use of implicit user feedback (e.g., information contained in billions of clicks, the structure of web links, etc.), in addition to explicit feedback from users and possibly some expert curation of answers to common questions.  Moreover, Google's system may decide which questions to display answers for, and probably limits itself to the answers that it is most confident in.  

Thus, the data in Google's answer boxes likely captures a variety of signals that contribute towards its high-quality. We believe aiming for a `standard' NLP QA system that's on par with Google QA is therefore a challenging and worthwhile goal.

\paragraph{Future uses of \name.}
One challenge in the progress on long-form QA is response evaluation. 
To facilitate future work on \name{} and replicability of our human evaluations, we have released the templates used for crowdsourcing human judgements.  
Efforts on text generation tasks such as ours will benefit from---and should in turn benefit advances in---proposals for streamlining human evaluation of models~\cite{khashabi2021genie}.

We hope our analysis and data will benefit the understanding of and further development of QA systems for dealing with diverse response types. 

While we used \name{} for the purposes of QA, 
we expect this data to have a variety of use-cases, such as building a \emph{knowledge-base} accessible via question queries~\cite{bosselut2019comet}, creating a better question generation system, etc. We leave such investigation to future work.


\section*{Acknowledgement}
The authors would like to thank Sihao Chen, Peter Clark, and Oyvind Tafjord for their help throughout this project. 
TPU machines for conducting experiments were provided by Google.

Chris Callison-Burch was supported in part by the DARPA KAIROS Program (contract
FA8750-19-2-1004), the DARPA LwLL Program (contract FA8750-19-2-0201), and the IARPA BETTER Program (contract 2019-19051600004). Hanna Hajishirzi was supported in part by  ONR N00014-18-1-2826, Allen Distinguished Investigator Award, and NSF CAREER award. Approved for Public Release, Distribution Unlimited. The views and conclusions contained herein are those of the authors and should not be interpreted as necessarily representing the official policies, either expressed or implied, of DARPA, IARPA, or the U.S. Government.

\bibliography{ref}
\bibliographystyle{acl_natbib}

\clearpage

\appendix

\label{sec:appendix}

\section{Query Terms}
\label{appendix:query:terms}
The  list of terms used for bootstrapping questions:  
``who'', ``whom'', ``whose'', ``what'', ``which'', ``when'', ``where'', ``why'', ``how'', ``should'', ``would'', ``wouldn't'', ``can'', ``can’t'', ``will'', ``won’t'', ``aren't'', ``do'', ``does'', ``has'', ``have'', ``am'', ``are'', ``is'', ``shouldn't'', ``isn't'', ``could'', ``couldn't'', ``does'', ``don’t'', ``must'', ``may'', ``ought''.

\section{Extracting Answers from Google}
\label{sec:answer:extraction}

For technical reasons, the answer extraction was done in two steps. 
(1) We first scrape the search results for all of our questions. 
This is the main extraction bottleneck as there is no official APIs to provide the answer boxes. Therefore, one needs to extract them directly from the HTML search results. 
We use Selenium\footnote{https://github.com/SeleniumHQ/selenium/} which simulates browser experience. 
Note one cannot send too many queries to Google in a short span of time (due to various query limits). 
Therefore, we ensured to have enough delays between our queries (otherwise, we'd be blocked). Overall, this extraction process was done in 3 months.  
Subsequent to extracting the search HTML results, (2) we extract answer strings from the HTML content of the search results. 
Answer types are also inferred at this stage, based on the HTML tags around the answer.


\section{Invalid Questions and Answers}
\label{appendix:invalids}
Based on the human evaluation of \name{} in \S\ref{subsec:data:quality}, we should example of erroneous instances.   Figure~\ref{fig:box1} shows examples of invalid questions. Often the questions are deemed invalid since they're under-defined or significantly deviate from the proper English. Figure~\ref{fig:box2} shows examples of invalid answers (to valid questions).  Invalid answers often do not sufficiently address the topic of the given question.

\begin{figure}[h]
\egbox{
    \fontsize{8pt}{10pt}\selectfont
    \textbf{Type:} curr-conv \\ 
    \textbf{Question:} 1 euro is hoeveel nok?  \\ 
    \textbf{Question:} how much is 45 in nigeria money? \\ 
    \noindent\rule[0.5ex]{\linewidth}{1pt}
    \textbf{Type:} time-conv \\
    \textbf{Question:} 2 am eastern standard time?  \\
    \textbf{Question:} what is the difference between china and republic of china?  \\
    \noindent\rule[0.5ex]{\linewidth}{1pt}
    \textbf{Type:} knowledge \\ 
    \textbf{Question:} what age is beauty and the beast? \\
    \textbf{Question:} acdc who made who live at donington?  \\ 
    \noindent\rule[0.5ex]{\linewidth}{1pt}
    \textbf{Type:} snippet \\ 
    \textbf{Question:} have mercy on me o god according to your loving kindness?  \\
    \textbf{Question:} dating a guy who is selfish? \\ 
    \noindent\rule[0.5ex]{\linewidth}{1pt}
    \textbf{Type:} collection \\ 
    \textbf{Question:} what are some areas of improvement?  \\
    \textbf{Question:} can sıkıntısına ne iyi gelir? 
}
\caption{examples of invalid questions}
\label{fig:box1}
\end{figure}

\begin{figure}[h]
\egbox{
    \fontsize{8pt}{10pt}\selectfont
    \textbf{Type:} time-conversion \\ 
    \textbf{Question:} what is the difference between mexican and spanish? \\ 
    \textbf{Answer:} Madrid, Spain is 7 hours ahead of Mexico City, CDMX, Mexico \\ 
    \noindent\rule[0.5ex]{\linewidth}{1pt}
    \fontsize{8pt}{10pt}\selectfont
    \textbf{Type:}  unit-conversion \\ 
    \textbf{Question:} what is 12 pm in spanish? \\ 
    \textbf{Answer:} 13:00 Saturday, in Madrid, Spain \\ 
    \noindent\rule[0.5ex]{\linewidth}{1pt}
    \fontsize{8pt}{10pt}\selectfont
    \textbf{Type:} snippet (short) \\ 
    \textbf{Question:} how many working days in january 2020 malaysia? \\ 
    \textbf{Answer:} 262 working days  \\ 
\noindent\rule[0.5ex]{\linewidth}{1pt}
    \fontsize{8pt}{10pt}\selectfont
    \textbf{Type:} knowledge \\ 
    \textbf{Question:} aids and hiv are acronyms for? \\ 
    \textbf{Answer:} HIV/AIDS \\
\noindent\rule[0.5ex]{\linewidth}{1pt}
    \textbf{Type:} snippet \\ 
    \textbf{Question:} are ralph lauren jackets good? \\ 
    \textbf{Answer:} Connoisseur. They are made by Corneliani in half-canvas construction. If you like them, they fit and you can afford them, they are a great value at 75\% off. I think they are still good at 50\% off.  \\ 
    \noindent\rule[0.5ex]{\linewidth}{1pt}
    \fontsize{8pt}{10pt}\selectfont
    \textbf{Type:} collection \\ 
    \textbf{Question:} how long does it take to get through john wayne airport? \\ 
    \textbf{Answer:} ['0-10 Min.', '11-20 Min.', '21+ Min.']
}
\caption{examples of invalid answers (to valid questions)}
\label{fig:box2}

\end{figure}

\clearpage

\section{Error Analysis of Model Predictions}
\label{subsec:error:analyais-appendix}

Here are examples of \emph{hallucination}, \emph{redundancy}, and \emph{incompleteness} errors in model predictions based on our human analysis discussed in \S\ref{subsec:error:analyais}. As noted earlier, all models exhibit very few \emph{incoherence} errors, mainly because this aspect is directly addressed during model pre-training.

\egbox{
    \fontsize{8pt}{10pt}\selectfont
    \textbf{Question:} how much does it cost to get court married?\\
    \textbf{Google ans:} Your application must be notarized. Please include at least one phone number where you can be reached during the day. Enclose a check or money order for \$69 payable to King County. The check or money order must include your address.  \\ 
    \textbf{Predicted ans:} The cost of a court marriage is between \$40,000 and \$50,000. The cost of a court marriage is between \$40,000 and \$50,000. The cost of a court marriage is between \$40,000 and \$50,000. \\ 
    \textbf{Explanation:} In this example, the model has made up  numbers that do not seem reasonable (hence, a `hallucination' error.) Additionally, it is repeating the same point (hence, a `redundancy' error.) 
}

\egbox{
    \fontsize{8pt}{10pt}\selectfont
    \textbf{Question:} who went to the superbowl last year? \\
    \textbf{Google ans:} Here's your answer: The Eagles won Super Bowl LII on Feb. 4, 2018, beating the Patriots 4133 to win the league championship. \\ 
    \textbf{Predicted ans:} The Super Bowl was a national championship game between the New York Giants and the New York Giants. The team was a member of the New York Giants, who starred in the Super Bowl in the 1970s. \\
    \textbf{Explanation:} In this example, there is repetition in the model's prediction (hence, a `redundancy' error.) 
}

\egbox{
    \fontsize{8pt}{10pt}\selectfont
    \textbf{Question:} what courses do you need to be a kinesiologist? \\
    \textbf{Google ans:} To become a kinesiologist, you typically need either a bachelor's or a master's degree in Kinesiology or a related field, such as Human Kinetics, Exercise Science, or Exercise Physiology. \\ 
    \textbf{Predicted ans:} To become a kinesiologist, you will need to complete a four-year degree in kinesiology, biology, chemistry, chemistry, physics. \\
    \textbf{Explanation:} In this example, the model provides some relevant points, but the response is not as detailed as the gold response (hence, an `incompleteness' error.)
}

\newpage

\section{Fine-grained Evaluation of Short-Answer Questions}
\label{subsec:short-answer-questions}
\begin{figure}[h]
    \centering
        \includegraphics[scale=0.7,trim=0.4cm 0.6cm -0.5cm 0cm]{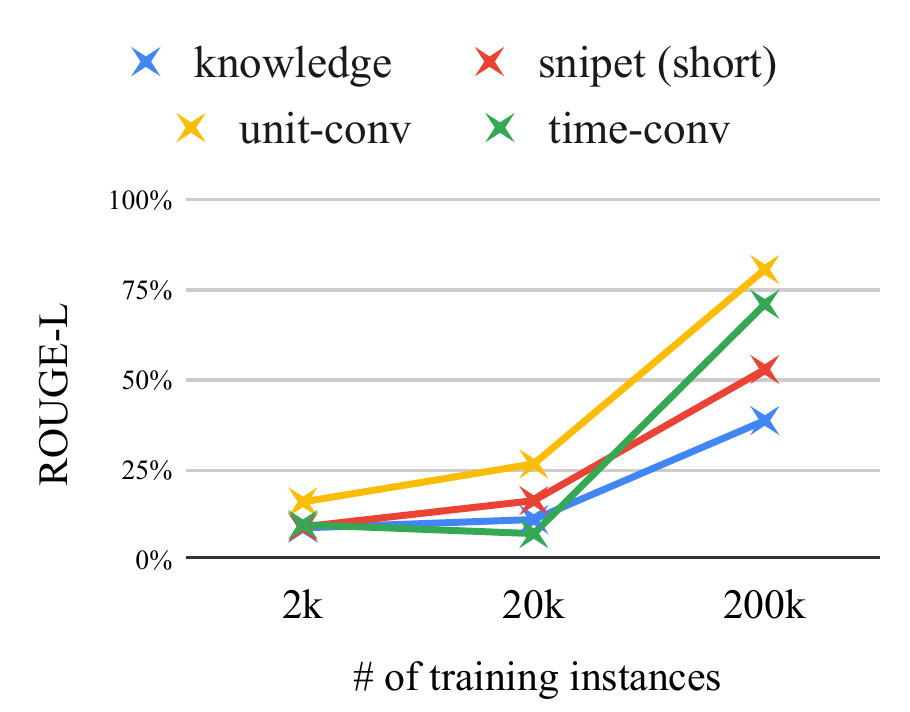}
        \includegraphics[scale=0.7,trim=0.4cm 0cm -0.5cm 1.6cm,clip=true]{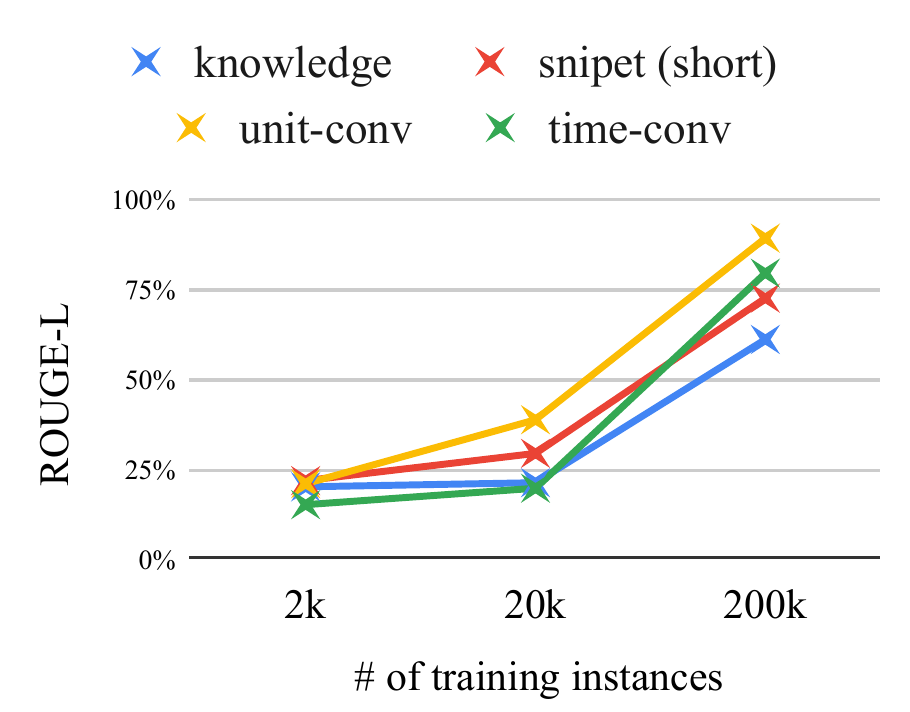}
    \vspace{-.2cm}
    \caption{
        \emph{Automatic} evaluation of T5 (small: top, 11B: bottom) models on different types of the questions included in short-answer sub-task (\task{short}). `unit-conversion' questions benefit the most from more labeled data, while `knowledge' lookup questions are the opposite. 
    }
    \label{fig:automatic:short}
\end{figure}

Fig.~\ref{fig:automatic:short} shows the breakdown of the automatic evaluation for different types of short response questions. As expected, certain niche question types (such as `unit-conversion') benefit the most from labeled data. 
In contrast, open-ended question types (such as `knowledge' lookup) benefit less from more labeled data.

\begin{figure*}[t]
    \centering
    \includegraphics[scale=0.55]{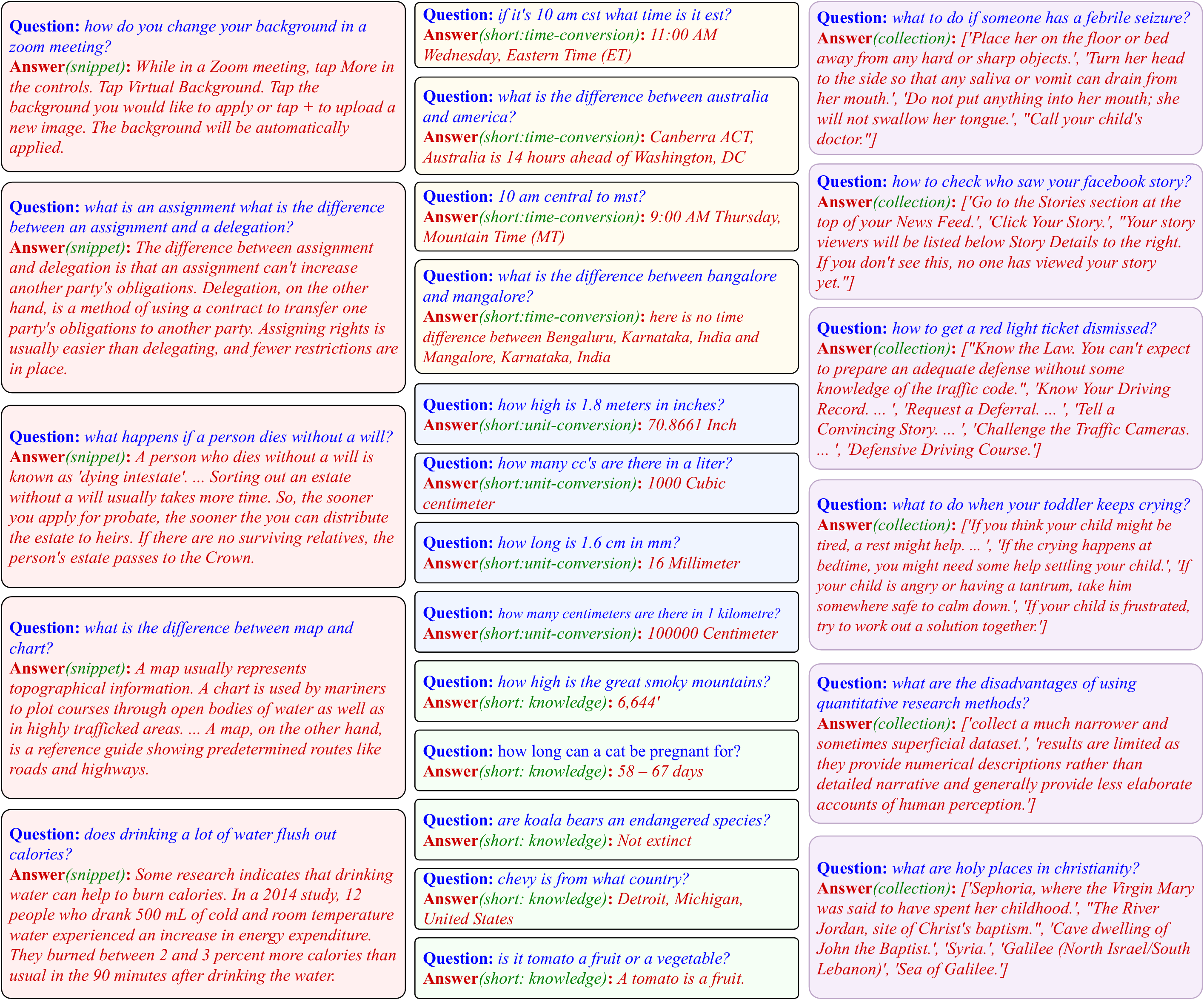}
    \caption{More examples from \name. Instances of questions with the same type share background colors. }
    \label{fig:more:examples}
\end{figure*}

\end{document}